\documentclass{bmvc2k}

\usepackage{multirow}
\usepackage{amsfonts}
\usepackage{xr}
\usepackage{cleveref}
\usepackage{mathrsfs}

\title{Few-shot Multispectral Segmentation with Representations Generated by Reinforcement Learning}

\addauthor{Dilith Jayakody}{dilith.18@cse.mrt.ac.lk}{1}
\addauthor{Thanuja Ambegoda}{thanuja@cse.mrt.ac.lk}{1}

\addinstitution{
 Department of Computer Science and Engineering\\
 University of Moratuwa\\
 Moratuwa, Sri Lanka
}

\runninghead{D. Jayakody \& T. Ambegoda}{IndexRL}

\begin{document}

\maketitle

\begin{abstract}
The task of segmentation of multispectral images, which are images with numerous channels or bands, each capturing a specific range of wavelengths of electromagnetic radiation, has been previously explored in contexts with large amounts of labeled data. However, these models tend not to generalize well to datasets of smaller size. In this paper, we propose a novel approach for improving few-shot segmentation performance on multispectral images using reinforcement learning to generate representations. These representations are generated as mathematical expressions between channels and are tailored to the specific class being segmented. Our methodology involves training an agent to identify the most informative expressions using a small dataset, which can include as few as a single labeled sample, updating the dataset using these expressions, and then using the updated dataset to perform segmentation. Due to the limited length of the expressions, the model receives useful representations without any added risk of overfitting. We evaluate the effectiveness of our approach on samples of several multispectral datasets and demonstrate its effectiveness in boosting the performance of segmentation algorithms in few-shot contexts. The code is available at \url{https://github.com/dilithjay/IndexRLSeg}.
\end{abstract}

%-------------------------------------------------------------------------
\section{Introduction}

Multispectral imagery is a powerful tool in a variety of applications in domains such as remote sensing, medical imaging, and thermal imaging. The inherent ability of multispectral images to capture data across various wavelengths of light provides information about the dynamics of the surface being captured. A fundamental task in capturing this information is image segmentation, which involves identifying distinct regions or objects based on certain criteria. However, existing work relies on large datasets to achieve good performance. The core challenge of working with smaller datasets is generalizing to a wider population.

Spectral indices have been widely employed (\cite{vss, msnet, riceseg}) as generalized representations of a class of interest. These indices are mathematical expressions between the bands of a multispectral image, designed to create representations based on the underlying reflective properties of the object of interest. For instance, the Normalized Difference Vegetation Index (NDVI) is commonly used to assess vegetation health, while the Normalized Difference Water Index (NDWI) is employed for water body detection.

However, the utility of spectral indices is constrained by their availability and adaptability to different contexts. Typically, these indices are designed for a limited set of predefined classes, making them less effective when applied to novel classes or datasets. Furthermore, the process of creating custom indices tailored to specific segmentation objectives is often a laborious and iterative trial-and-error procedure that demands substantial domain expertise.

\begin{figure}
  \includegraphics[width=\columnwidth]{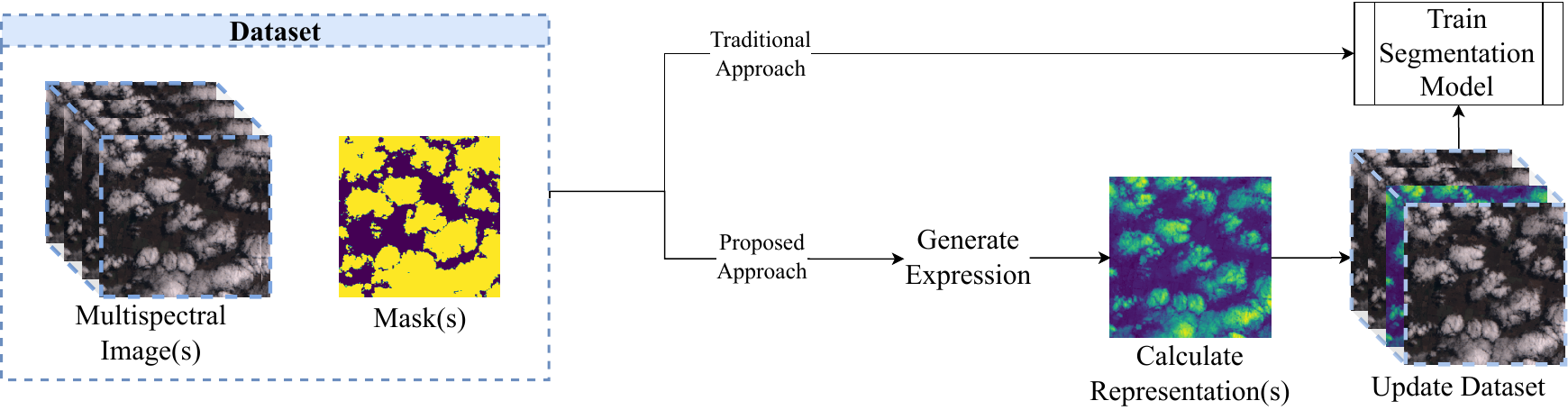}
  \caption{A one-shot example of the proposed approach.}
  \label{fig:1}
  \centering
\end{figure}

In this paper, we propose a novel approach for improving segmentation performance on multispectral images, as demonstrated by Figure \ref{fig:1}. Our system begins by discovering a mathematical expression (in other words, a spectral index) that is expected to yield a good segmentation performance. We accomplish this using a reinforcement learning formulation to explore possible expressions/indices, train a model to generate expressions with higher rewards, and then identify a new set of expressions with the improved model to facilitate better exploration. After identifying a suitable mathematical expression, we evaluate the expression on each image and integrate the resulting channel with the rest (or a subset of the rest) of the channels of the image. In other words, the proposed approach can be interpreted as a data augmentation technique that helps to cope with the lack of data. Furthermore, due to the finite number of variations that the mathematical expression could take, this approach enables the inclusion of useful representations without an added risk of overfitting, a common pitfall when training on small datasets.

% The reasoning behind the hypothesis that spectral indices would help improve segmentation performance is that certain mathematical functions are difficult for neural networks to represent in a few-shot context. For example, a simple multiplication between two channels would require a large number of training samples in addition to a sufficiently large network to be accurate over a wide range of inputs. By using spectral indices as input features, we can provide the neural network with more informative and relevant input data, improving its ability to learn and generalize to new examples. Empirically, this hypothesis holds.

Accordingly, the contributions of our research are as follows:
\begin{itemize}
    \item We propose a novel application of reinforcement learning for discovering spectral indices that improve few-shot segmentation performance in multispectral images.
    \item We address the challenging task of few-shot multispectral segmentation which, to the best of our knowledge, has not been previously explored in the literature in a general context.
    \item We demonstrate the effectiveness of our proposed approach on several datasets by comparing the performance against several baseline models.
\end{itemize}

\section{Background and Related Work}

\paragraph{Spectral indices for segmentation.} Spectral indices have been used to assist segmentation in a variety of techniques. Traditional methods directly use algorithms such as Otsu's thresholding \cite{otsu} or watershed algorithms \cite{watershed} to binary segment the single channel result of evaluating spectral indices \cite{riceseg, vegidxseg, vss} (see Section \ref{sec:index-eval} for more details on evaluating indices). However, these techniques are only viable when there already exist spectral indices tailored to the classes of interest. Our work draws inspiration from MSNet \cite{msnet} which utilizes spectral indices for the segmentation of arbitrary classes. The authors demonstrate the benefits of incorporating the Normalized Difference Vegetation Index (NDVI) and the Normalized Difference Water Index (NDWI) as additional channels of multispectral images for segmentation. This suggests that these indices carry some information that is not easily represented by deep learning models.

\vspace{-0.3cm}
\paragraph{Automated remote sensing index generation.} In \cite{deepindices}, Vayssade et al. explore index generation under a finite set of predefined forms of equations (such as linear and linear ratio) and achieve promising results in several vegetation classes. In contrast, our approach does not limit the form of the generated expressions and explores classes in a broader set of domains.

\vspace{-0.3cm}
\paragraph{RL for expression induction and theorem proving.} The tasks of expression induction and theorem proving follow a formulation similar to that of index generation as explored in this research. One early instance of these tasks is found in \cite{symb-exp} where an RL agent is employed to generate symbolic expressions, specifically polynomial expressions. In this setup, an agent, implemented as a Recurrent Neural Network (RNN), iteratively selects symbols that collectively form an expression in post-order notation. Building upon a similar foundation, \cite{symb-pol} adopts an analogous strategy to leverage RL to estimate the policy function of an RL agent, with the dimensions of the state serving as the operands of the mathematical expression. RL-based agents have also been used to navigate tableaux trees (trees with branches representing sub-formulae of theorems) for proving first-order logic \cite{proof-18, proof-21}. The representation of the state in our methodology contrasts with that of these techniques. As opposed to using a tree-based representation with post-order traversal, we use the expression symbols under in-order traversal to represent the state. This makes each state directly human-readable, resulting in an improved explainability of a given state.

% \begin{figure}
%   \includegraphics[width=\textwidth]{figures/index-generator.pdf}
%   \caption{\textbf{Index Generator Training Process.} Each training iteration of the index generator has 2 main phases: the data collection phase and the policy training phase. Expression samples are generated during the data collection phase by a generative model that guides MCTS. The model is improved by using the generated samples to train the policy network during the policy training phase.}
%   \label{fig:index-generator}
%   \centering
% \end{figure}

\vspace{-0.3cm}
\paragraph{Monte Carlo Tree Search (MCTS).} MCTS \cite{mcts} is an algorithm that uses simulations to determine the best action to be taken from a given state. The simulation is done iteratively where each iteration consists of selecting an action, expanding a branch on a tree for the selected action, followed by the simulation from a leaf node on the tree. The exploration-exploitation trade-off is managed by a measure (often the Upper Confidence Bound (UCB1) score \cite{ucb}) which prioritizes nodes that have a low visitation count and high returns after simulation. While the vanilla MCTS algorithm takes random actions during simulation, later approaches \cite{alphago, expert-iter, alphazero, alphatensor} use a neural network to guide the simulation. These algorithms are broadly known as Neural MCTS. Our proposed methodology builds on these concepts by using a GPT-based model architecture as the policy network to guide the simulation in identifying an effective spectral index for a given segmentation task.

\vspace{-0.3cm}
\paragraph{Few-shot Segmentation.} Few-shot segmentation is the task of segmenting objects using a small amount of labeled data. Most existing work approaches this task by comparing the unlabeled image (query image) against the labeled image (support image) at inference time \cite{fss}. However, some work uses techniques such as data augmentation \cite{med-aug, deform, spot} and sample synthesis \cite{laso, hallu}, either as standalone techniques or for boosting the performance of existing techniques. Our solution falls into the latter category of performance-boosting algorithms, as it can be considered a preprocessing step.

\section{Methodology}
\begin{figure}
  \includegraphics[width=\columnwidth]{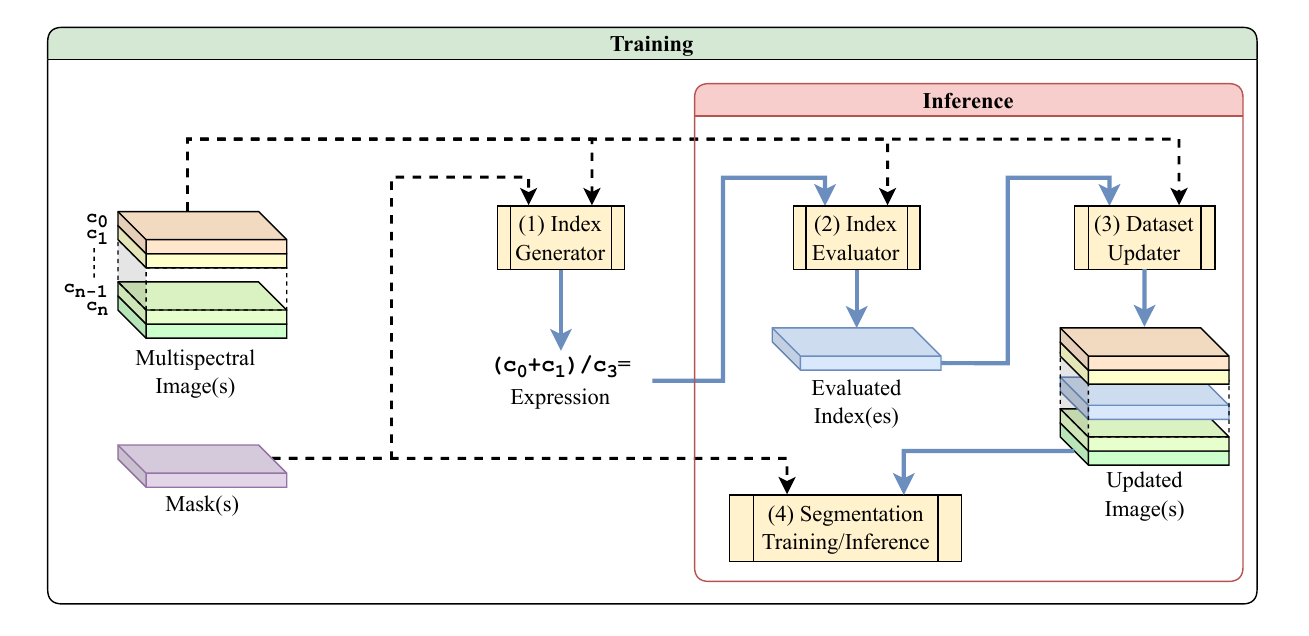}
  \caption{
  The proposed methodology and its four main components. (1) Index Generator (2) Index Evaluator (3) Dataset Updater (4) Segmentation Trainer.}
  \label{fig:methodology_overview}
  \centering
\end{figure}

Figure \ref{fig:methodology_overview} presents the proposed methodology consisting of four main components: (1) Index Generator: Uses the train set (image-mask pairs) to generate an expression. (2) Index Evaluator: Uses the generated expression and the original images to create the evaluated indices. (3) Dataset Updater: Uses the evaluated indices to update the original images. (4) Segmentation Trainer: Trains a segmentation model using the updated images.

We may interpret this process as follows. The first component identifies the best augmentation for the data. The second component executes the augmentation on each input image to create its respective single-channel augmented image. The third component combines each augmented channel with its respective original input image. In the sections that follow, we discuss the first three components. The fourth component, the segmentation trainer, can be any multispectral segmentation approach.

\subsection{Index Generator}
\label{sec:index-gen}

% As shown in Figure \ref{fig:index-generator}, the approach takes the labeled dataset as input, generates a spectral index based on the input, updates the labeled dataset using the generated index, trains a segmentation model using the updated dataset, and returns the trained model and the spectral index as outputs.

% At inference time, unlabeled data is updated using the remote sensing index and evaluated on the trained model.

We begin this section by describing the formulation of the RL problem. This will be followed by a description of the agent's training process.

\subsubsection{RL Formulation}

% Formulating the problem of spectral index generation as a reinforcement learning task can be done by defining the state space, action space, environment, and reward function.

\paragraph{States and Actions.}
The \textbf{state} is defined by the currently generated portion of the mathematical expression while the \textbf{actions} are represented by the symbols that form the expression. Additionally, a terminal state is defined by the presence of the "=" symbol at the end of the expression. Accordingly, the action space consists of the following symbols:
\begin{itemize}
  \item \textit{Expression symbols}: (, ), +, -, *, /, square, square root (8 symbols)
  \item \textit{Image Channels}: $c_0$, $c_1$, ... , $c_{n_{channels} - 1}$ ($n_{channels}$ symbols)
  \item \textit{End symbol:} $=$ (1 symbol)
\end{itemize}

At each step, the environment appends the newly generated action/symbol to the current expression and returns it as the output state, along with the reward calculated using the reward function. The maximum length of the expression/state is a hyperparameter that can be adjusted based on the context or domain (longer expressions create more complex representations).

\phantomsection
\label{sec:reward}
\paragraph{MCTS Reward.}
The reward for guiding MCTS is defined for three different cases as follows.

\begin{equation}
R(s) = \begin{cases}
    -1, & \text{if episode ended at an invalid state}\\
    r(s), & \text{if episode ended at a valid state}\\
    0,  & \text{otherwise (episode has not ended)}\\
\end{cases}
\end{equation}

Here, the function $r(s)$ approximates how good the given state $s$ is for improving the segmentation performance. In addition to the expression, the reward function shall also use the existing labeled images to calculate the reward. An evaluated index will be calculated for each image of the training set as per Section \ref{sec:index-eval}, which can thereby be used to calculate the reward. Each evaluated index is a single-channel image with the same height and width as the original image. As part of this research, we evaluate several different reward functions as choices for $r(s)$ that follow the same base structure shown below:

\begin{align}
    score & = f(\mathcal{E}, M) \label{eq:score} \\
    score' & = f(\mathcal{J} - \mathcal{E}, M) \label{eq:score_inv} \\
    r(s) & = min\{score, score'\} \label{eq:r_s}
\end{align}

\noindent
where $\mathcal{E} \in \mathbb{R}^{H \times W}$ refers to an evaluated index that has been preprocessed (see Appendix \ref{supp-sec:index-proc} for details), $M \in \mathbb{R}^{H \times W}$ refers to the ground truth segmentation mask corresponding to the image from which $\mathcal{E}$ is produced, $f(\mathcal{E}, M)$ refers to the heuristic function for estimating the utility of using $\mathcal{E}$ to predict $M$, and $\mathcal{J} \in \mathbb{R}^{H \times W}$ is a matrix of ones. Note that the final score for a given expression is the average of $r(s)$ over all or a subset of the training images.

The evaluated functions for the choice of $f$ in equations \ref{eq:score} and \ref{eq:score_inv} include the F1 Score (with a threshold for $E$ of 0.5 to get a binary mask), Area Under the Curve (AUC) (with threshold, similar to the F1 score), Cosine Similarity (CS), Intersection over Union (IoU), and the Pearson Correlation Coefficient (PCC).

The intuition behind the use of the minimum of $score$ and $score'$ is that $\mathcal{J} - \mathcal{E}$ is a simple operation for any network, given $\mathcal{E}$, and either of the two may be the default representation being used within the network, based on the initialization of weights. Obtaining the minimum of the two scores penalizes the lack of information in either of the representations.

\phantomsection
\label{sec:training-reward}
\paragraph{Training Reward.}
Once an expression is generated, we calculate the actual segmentation performance of the expression by training a model using only the evaluated indices. Since the expressions used in training are generated much less frequently than in MCTS simulations, this is a practical way to obtain a more accurate score for the performance gained from the expression.

\subsubsection{RL Training Process}
\label{sec:rl-training}

% \begin{figure}[ht]
%   \centering
%   \includegraphics[width=\linewidth]{process.pdf}
%   \caption{RL Training Process overview}
% \end{figure}

\begin{figure}
  \includegraphics[width=\textwidth]{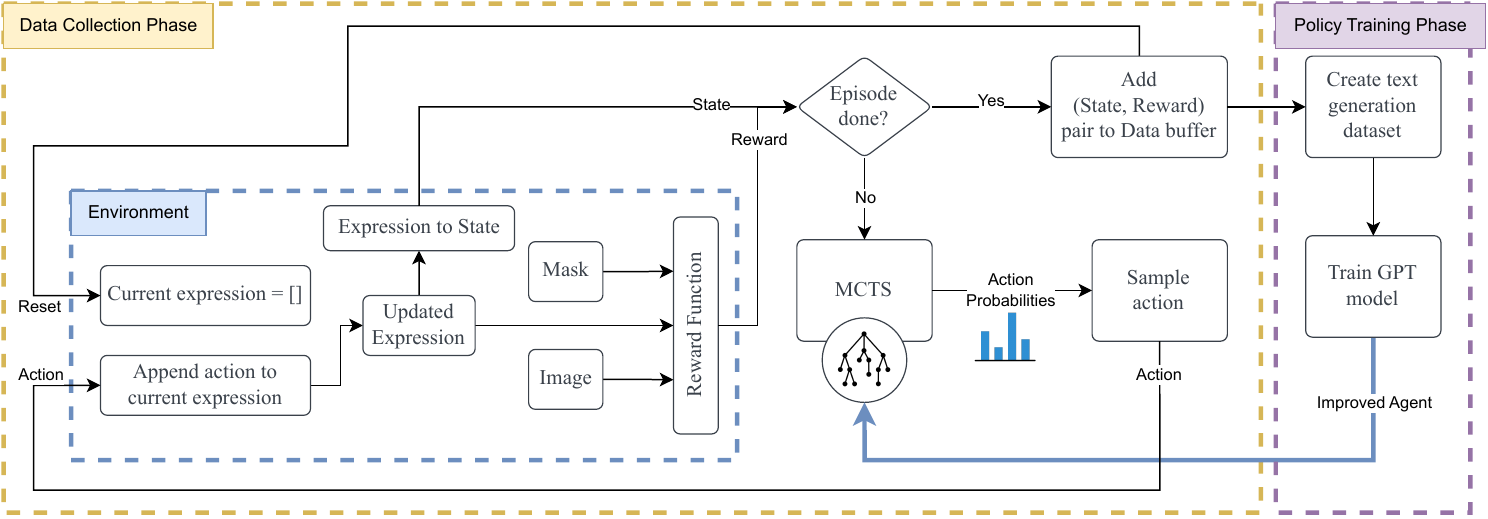}
  \caption{\textbf{Index Generator Training Process.} Each training iteration of the index generator has 2 main phases: the data collection phase and the policy training phase. Expression samples are generated during the data collection phase by a generative model that guides MCTS. The model is improved by using the generated samples to train the policy network during the policy training phase.}
  \label{fig:index-generator}
  \centering
\end{figure}

First, the agent's policy network is pretrained to generate valid expressions as output. The specifics of this pretraining can be found in Appendix \ref{supp-sec:adaptations}. The training process that follows requires as input a set of multispectral images, with segmentation masks for each image. Each iteration of the training process is performed in two phases: (1) Data Collection and (2) Policy training. Figure \ref{fig:index-generator} shows the flow of steps in each phase and iteration, and the paragraphs below discuss this further.

\paragraph{Data Collection Phase.}
Each iteration uses MCTS to generate a probability distribution over the actions based on the simulated return observed for each action. This probability distribution is then used to sample the next action to be taken by the environment. If the action results in the termination of the episode (in other words, if the last action is the “=” symbol), then the expression and its training reward (as described in Section \ref{sec:training-reward}) are passed into the data buffer. Otherwise, if the episode continues, the state is passed into MCTS for simulation. Each iteration of the data collection phase generates several expressions. The number of generated expressions is a tunable hyperparameter (we use 10 for our experiments). The generated expressions are stored in a data buffer. Of the stored expressions, the expressions that yielded a high reward are passed for training.
% The criteria for classifying an expression as "high-reward" is discussed in the \nameref{section:adaptive-buffer} section.

% While one image and its mask are sufficient to execute the data collection phase, the result is likely to be more generally beneficial if the average of more images is used to calculate the reward. Thus, there is a configurable trade-off between the lower execution time and the discovery of the better-performing expression.

\paragraph{Policy Training Phase.}
In the policy training phase, the policy network, which follows a GPT-based model architecture (the model configuration is available in Appendix \ref{supp-sec:gpt-config}), is trained to generate useful expressions. In other words, given the current state, we train the model to generate the next symbol. With each iteration, the model learns to generate better-performing expressions. Subsequently, the MCTS exploration improves. This is because MCTS only explores a limited number of branches, and the exploration thereafter is simulated using the symbols generated by the model. A better-performing policy network can explore higher-reward expressions without the guidance of MCTS.

% \subsection{Using the expression in segmentation}
% During the training of the reinforcement learning agent, the agent is expected to explore a number of indices/expressions. Accordingly, the index that gave the best performance can be picked and used for training a segmentation model. This is performed by evaluating the expression for each image of the dataset and concatenating the evaluated indices with their respective images.

% The hypothesis is that training the segmentation model on these images will yield better performance compared to a regular segmentation approach.

\subsection{Index Evaluator}
\label{sec:index-eval}
The index evaluator uses the generated expression to create a single-channel image that we refer to as the \textbf{evaluated index}. This is obtained by performing pixel-wise operations across the channels based on the expression.

For example, say the chosen expression is $c0 \times c1 =$. To evaluate this expression, for each image, we perform an element-wise multiplication between the pixels of the channel at index 0 and those at index 1.

\vspace{-0.1cm}
\subsection{Dataset Updater}
\label{sec:dataset_updater}
We explore two main techniques by which the dataset can be updated using the evaluated index.

\vspace{-0.4cm}
\paragraph{Concatenation.} The first integration mode simply concatenates each evaluated index with its respective source image. So, given a 10-channel image, the concatenation would result in an 11-channel image.

\vspace{-0.5cm}
\paragraph{Replacement.} The second technique is to replace a channel with an evaluated index. The evaluated index is only substituted with channels that appear within the expression to avoid the loss of information.

\vspace{-0.3cm}
\paragraph{}Based on the above two techniques, we define four modes: Single-index Concatenation (\textbf{C}), Multiple-index Concatenation (\textbf{CM}), Single-index Replacement (\textbf{R}), and Multiple-index Replacement (\textbf{RM}). Each mode is empirically evaluated in Section \ref{sec:ablation}. The \textbf{best updating mode} for a given dataset is determined by observing the validation accuracy of each mode.

\section{Experiments}

\paragraph{Datasets.}
Due to the absence of previously created few-shot multispectral segmentation datasets, we explore the performance of the proposed approach across several multispectral datasets, MFNet \cite{mfnet}, Sentinel-2 Cloud Mask Catalogue \cite{cloud}, Landslide4Sense \cite{landslide}, and RIT-18 \cite{rit-18}. For MFNet and RIT-18, which are multiclass segmentation datasets, we evaluate our approach on selected classes (car, person, and bike on MFNet; grass and sand on RIT-18). The datasets containing samples of each class are treated as individual datasets during experimentation.  Accordingly, the final set of datasets consists of car, person, bike, cloud, landslide, grass, and sand. To accommodate a few-shot context, we chose to randomly sample 50 data points from each dataset. These 50 data points are split into train, validation, and test in the frequencies of 20, 10, and 20, respectively, with each image being of shape $N_{channels} \times 256 \times 256$.

% \vspace{-0.3cm}
\subsection{Index Generation Experimental Details}

\paragraph{Reward Heuristic Comparison.}
As mentioned in Section \ref{sec:reward}, we evaluate the performance of several heuristic functions to estimate the segmentation performance for a given expression. We sample 200 randomly generated expressions and each sampled expression is assigned a score by obtaining the actual segmentation performance (IoU) through training on the training sets of the MFNet datasets. Next, we calculate the heuristic scores for each of the expressions as defined by equation \ref{eq:r_s}. Finally, we perform a correlation analysis between the scores of each heuristic function and the scores obtained through training, by calculating the t-statistic.

Based on this analysis (see Appendix Table \ref{supp-tab:heuristic-comparison}), we observe that for the evaluated dataset, only the Pearson correlation (PCC) shows a statistically significant correlation with the training performance, at a significance level of 0.05. Thus, the PCC is used as the reward heuristic to evaluate each branch of the MCTS.

% \vspace{-0.2cm}
\paragraph{Index Selection.}
As stated in Section \ref{sec:rl-training}, the data buffer used in the RL algorithm contains the best-performing expressions at any given time. After allowing the RL algorithm to explore for a satisfactory amount of time, we choose the two best expressions for the segmentation experiments.

% \vspace{-0.2cm}
\subsection{Segmentation Experimental Details}

We evaluate the performance of the proposed approach on three segmentation model architectures, UNet \cite{unet}, UNet++ \cite{unet++}, and DeepLabV3 \cite{deeplabv3}. For each model, we use a ResNet50 model \cite{resnet} pretrained on ImageNet \cite{imagenet} as the encoder. During training, all but the decoder of the model is frozen. We use the AdamW optimizer \cite{adamw} with beta coefficients of 0.9 and 0.999, epsilon of 1e-8, and a weight decay of 1e-2. Each model is trained with learning rates of 1e-3 and 1e-4. The models undergo early stopping with 1000 epochs of patience by monitoring the validation loss. The trained models are then evaluated on the test set to obtain the final performances. We keep all the above factors the same when comparing against the baseline, the only difference being that the dataset fed into the baseline has not been integrated with a remote sensing index using a dataset updating mode.

\subsection{Overall Results}

\begin{table*}[t]
  \centering
  \begin{tabular}{l|ll|ll|ll}
    \hline
    \multirow{2}{*}{Dataset} & \multicolumn{2}{c|}{UNet} & \multicolumn{2}{c|}{DeepLabV3} & \multicolumn{2}{c}{UNet++} \\
    \cline{2-7}
     & Baseline & Ours & Baseline & Ours & Baseline & Ours \\
    \hline
    car 	& 62.5 	& \textbf{67.4} (RM)	& 55.4 	& \textbf{58.2} (RM)	& \textbf{74.8} 	& \textbf{74.8} (CM) \\
    person 	& 46.4 	& \textbf{48.4} (RM)	& 17.9 	& \textbf{25.2} (RM)	& 47.3 	& \textbf{48.5} (R) \\
    bike 	& 37.8 	& \textbf{39.8} (RM)	& 31.1 	& \textbf{53.5} (RM)	& \textbf{40.3} 	& 36.4 (R) \\
    cloud 	& 80.6 	& \textbf{83.3} (RM)	& 62.0 	& \textbf{65.6} (R)	& 82.3 	& \textbf{84.2} (RM) \\
    landslide 	& 38.0 	& \textbf{43.1} (RM)	& 18.7 	& \textbf{20.5} (R)	& 35.9 	& \textbf{42.8} (RM) \\
    grass 	& 58.0 	& \textbf{73.7} (RM)	& \textbf{66.6} 	& 65.6 (R)	& 60.9 	& \textbf{70.3} (RM) \\
    sand 	& 13.1 	& \textbf{59.4} (RM)	& 12.6 	& \textbf{41.3} (RM)	& 25.2 	& \textbf{69.2} (RM) \\
    \hline
  \end{tabular}
  \caption{\textbf{Overall Results}. The table compares the IoU scores of the \textbf{baseline} model against that of the \textbf{best} dataset updating mode for each dataset class when trained on each model. (See Section \ref{sec:dataset_updater})}
  \label{tab:overall_results}
\end{table*}

\paragraph{}We compare the performance of the overall method against each baseline model and dataset (Table \ref{tab:overall_results}). While the proposed method results in a significant improvement in UNet, it can be observed that this advantage decreases slightly with increasing model size. We hypothesize that bigger models depend relatively less on the input representations. A qualitative analysis of the results on UNet can be found in Appendix \ref{supp-sec:qualitative}.

\subsection{Ablation Studies}
\label{sec:ablation}

\begin{table*}[t]
  \centering
  \fontsize{10pt}{20pt}
  \begin{tabular}{l|lllll|lllll}
    \hline
    \multirow{2}{*}{Dataset} & \multicolumn{5}{c|}{UNet} & \multicolumn{5}{c}{UNet++} \\
    \cline{2-11}
     & B & C & CM & R & RM & B & C & CM & R & RM     \\
    \hline
    car 	& 62.5	& 63.0	& 61.1	& 65.0	& \textbf{67.4}	& \textbf{74.8}	& 72.0	& \textbf{74.8}	& 60.6	& 71.2 \\
    person 	& 46.4	& 41.8	& 47.0	& 45.1	& \textbf{48.4}	& 47.3	& 51.2	& \textbf{53.5}	& 48.5	& 45.1 \\
    bike 	& 37.8	& 29.3	& 37.4	& 37.0	& \textbf{39.8}	& 40.3	& 38.6	& \textbf{43.0}	& 36.4	& 33.7 \\
    cloud 	& 80.6	& 82.7	& 82.0	& 79.9	& \textbf{83.3}	& 82.3	& 83.0	& 81.9	& 83.9	& \textbf{84.2} \\
    landslide 	& 38.0	& 36.7	& 37.1	& 33.7	& \textbf{43.1}	& 35.9	& 36.8	& 33.5	& 34.3	& \textbf{42.8} \\
    grass 	& 58.0	& 61.7	& 63.3	& 73.6  & \textbf{73.7}	& 60.9	& 61.7	& 60.4	& \textbf{73.6}	& 70.3 \\
    sand 	& 13.1	& 12.3	& 11.5	& 44.3	& \textbf{59.4}	& 25.2	& 25.2	& 37.8	& 62.8	& \textbf{69.2} \\
    \hline
  \end{tabular}
  \caption{\textbf{Effects of Dataset Updating Mode.} The table compares the IoU scores of the \textbf{baseline} model against each updating mode. While RM can be thought of as a safe updating mode in most cases, it seems to be of more benefit to smaller models (UNet) in contrast to larger models (UNet++).}
  \label{tab:update_mode}
\end{table*}

\paragraph{Effects of Dataset Updating Mode.}
We evaluate the effects of the different dataset updating modes discussed in Section \ref{sec:dataset_updater} across all seven datasets and three models of interest (Table \ref{tab:update_mode}). It can be observed that in most cases, updating the dataset through some mode leads to an improvement in performance. As for which mode to use for updating in practice, RM is shown to be a safe first choice to evaluate with. However, this advantage seems to be less dominant in the case of UNet++. We hypothesize that the larger model size of UNet++ makes it less dependent on the input representations. A performance comparison of updating the grass dataset using NDVI as opposed to the generated index can be found in Appendix \ref{supp-sec:ndvi}.

\begin{table}
  \centering
  \begin{tabular}{l|ll|ll|l}
    \hline
    \multirow{2}{*}{Size}   & \multicolumn{2}{c|}{UNet} & \multicolumn{2}{c|}{UNet++} & Mean    \\
    \cline{2-5}
                            & B & RM & B & RM & increase \\
    \hline
    1   & 22.4 & \textbf{35.0} & 22.1 & \textbf{36.1} & +13.3 \\
    5   & 30.3 & \textbf{38.8} & 33.5 & \textbf{41.6} & +8.3 \\
    20  & 38.0 & \textbf{43.1} & 35.9 & \textbf{42.8} & +6.0 \\
    40  & 40.6 & \textbf{47.9} & 41.1 & \textbf{50.8} & +8.5\\
    80  & 47.4 & \textbf{50.0} & 46.5 & \textbf{48.2} & +2.2\\
    160 & 49.9 & \textbf{52.7} & 49.5 & \textbf{50.5} & +1.9\\
    \hline
  \end{tabular}
  \caption{\textbf{Effects at Different Training Set Sizes}. The table shows the comparison between the IoU scores of the multiple-index replacement method (RM) and the baseline (B) across training samples of different sizes from the Landslide4Sense dataset \cite{landslide}.}
  \label{tab:train_size}
\end{table}

\paragraph{Effects at Different Training Set Sizes.}
We evaluate the effects of the proposed methodology when the training set size is 1, 5, 20, 40, 80, or 160 samples (Table \ref{tab:train_size}). The Landslide4Sense \cite{landslide} is chosen for this evaluation due to the relatively lower performance scores across all experiments, the reason being that it leaves more room for improvement. Using Multiple-Index Replacement (RM) as the choice of the dataset updating mode, it can be observed that while the overall performance improves with increasing training set size, the benefits of using the indices tend to be higher for smaller training set sizes. We hypothesize that this is because the indices provide a more generalizable representation that may be lacking in smaller training sets.

\begin{table}
  \centering
  \begin{tabular}{l|ll|ll|ll|ll}
    \hline
    \multirow{2}{*}{Dataset}   & \multicolumn{2}{c|}{UNet} & \multicolumn{2}{c|}{UNet++}  & \multicolumn{2}{c|}{MSNet} & \multicolumn{2}{c}{CAINet}    \\
    \cline{2-9}
                            & B & RM & B & RM & B & RM & B & RM \\
    \hline
    MFNet   & 79.6 & \textbf{81.2} & 77.7 & \textbf{79.3} & 78.8 & \textbf{79.5} & 85.1 & \textbf{85.8} \\
    RIT-18  & 66.6 & \textbf{67.5} & 83.1 & \textbf{85.3} & 70.8 & \textbf{78.9} & 58.3 & \textbf{82.4} \\
    \hline
  \end{tabular}
  \caption{\textbf{Effects of Generated Indices on Multiclass Segmentation}. The table compares the IoU scores of the multiple-index replacement method (RM) and the baseline (B) across two multiclass segmentation datasets and four segmentation model architectures (two regular segmentation architectures: UNet \cite{unet}, UNet++ \cite{unet++}, and two multispectral segmentation architectures: MSNet \cite{msnet}, CAINet \cite{cainet}).}
  \label{tab:multiclass}
\end{table}

\paragraph{Effects of Generated Indices on Multiclass Segmentation.}
We evaluate how an index generated for a specific class can affect the performance of multiclass segmentation. We perform this evaluation on the MFNet dataset \cite{mfnet} and the RIT-18 dataset \cite{rit-18} by comparing the baseline training performance (B) against the performance observed through Multiple-Index Replacement (RM). For RM, we use the indices generated for the "car" class to update the MFNet dataset and those generated for the "grass" class to update the RIT-18 dataset. The results, as shown in Table \ref{tab:multiclass}, demonstrate that the indices contribute to the overall segmentation performance in the multiclass context as well. In addition, it should be noted that even models developed for the RGB-T context (such as CAINet \cite{cainet}) not only seem to gain performance improvements under the proposed method but also demonstrate potential in supporting other contexts (as with the RIT-18 dataset).

% \begin{table}
%   \centering
%   \begin{tabular}{l|l|ll|ll}
%     \hline
%     \multirow{2}{*}{Size}   & \multirow{2}{*}{Baseline}   &  \multicolumn{2}{c|}{NDVI} & \multicolumn{2}{c}{Generated}   \\
%     \cline{3-6}
%                            & & R & RM & R & RM \\
%     \hline
%     UNet    & 58.0 & 74.0 & 73.6 & 75.1 & 72.0 \\
%     UNet++  & 60.9 & 70.2 & 72.4 & 73.5 & 75.3 \\
%     \hline
%   \end{tabular}
%   \caption{\textbf{Effects of Generated Indices versus Existing Indices}. The table shows the comparison between the IoU scores of the multiple-index replacement method (RM) and the single-index replacement method (R) for the NDVI index and the index generated by the proposed method for the Grass class of the RIT-18 dataset \cite{rit-18}.}
%   \label{tab:ndvi}
% \end{table}

% \paragraph{Effects of Generated Indices versus Existing Indices.} To evaluate the performance of the generated indices in comparison to pre-existing indices, we compare NDVI to the expression for the Grass class of the RIT-18 dataset. We choose this setting due to the utility of NDVI in the identification of greenery in prior work.

\section{Conclusion}

% In this paper, we presented an approach for improving few-shot multispectral segmentation performance by generating mathematical expressions tailored to a class of interest using reinforcement learning. Each generated expression defines a data augmentation using which a single-channel image (a.k.a. an evaluated index) is created corresponding to each input image. The results demonstrate that replacing multiple channels of the input image with such evaluated indices from multiple expressions tends to lead to the best performance improvement.

In this paper, we presented an approach for improving few-shot multispectral segmentation performance. We achieve this by using reinforcement learning on a few labeled samples to generate expressions that define data augmentations. Each generated expression is used to augment each input image into a single-channel image (a.k.a. an evaluated index). The results demonstrate that replacing multiple channels of the input image with such evaluated indices from multiple expressions tends to lead to the best performance improvement.

The proposed algorithm is, however, limited by the amount of time it takes to generate a suitable index. Despite this limitation, once an index is generated, it can be used for any subsequent task pertaining to the dataset. Additionally, the current reward function is primarily targeted for binary segmentation. However, we demonstrate that multiclass segmentation can still benefit from the generated indices (Table \ref{tab:multiclass}). Future work can explore reward functions that directly target multiclass segmentation and the cross-compatibility of the generated expressions with other computer vision tasks such as image classification and object detection. Furthermore, in addition to the context of computer vision, the proposed RL formulation may also be used in traditional machine learning contexts to perform feature engineering and feature extraction to maximize performance.

\bibliography{egbib}

\begin{thebibliography}{34}
\providecommand{\natexlab}[1]{#1}
\providecommand{\url}[1]{\texttt{#1}}
\expandafter\ifx\csname urlstyle\endcsname\relax
  \providecommand{\doi}[1]{doi: #1}\else
  \providecommand{\doi}{doi: \begingroup \urlstyle{rm}\Url}\fi

\bibitem[Alfassy et~al.(2019)Alfassy, Karlinsky, Aides, Shtok, Harary, Feris, Giryes, and Bronstein]{laso}
Amit Alfassy, Leonid Karlinsky, Amit Aides, Joseph Shtok, Sivan Harary, Rogerio Feris, Raja Giryes, and Alex~M Bronstein.
\newblock Laso: Label-set operations networks for multi-label few-shot learning.
\newblock In \emph{Proceedings of the IEEE/CVF conference on computer vision and pattern recognition}, pages 6548--6557, 2019.

\bibitem[Anthony et~al.(2017)Anthony, Tian, and Barber]{expert-iter}
Thomas Anthony, Zheng Tian, and David Barber.
\newblock Thinking fast and slow with deep learning and tree search.
\newblock \emph{Advances in neural information processing systems}, 30, 2017.

\bibitem[Auer et~al.(2002)Auer, Cesa-Bianchi, and Fischer]{ucb}
Peter Auer, Nicolo Cesa-Bianchi, and Paul Fischer.
\newblock Finite-time analysis of the multiarmed bandit problem.
\newblock \emph{Machine learning}, 47:\penalty0 235--256, 2002.

\bibitem[Catalano and Matteucci(2023)]{fss}
Nico Catalano and Matteo Matteucci.
\newblock Few shot semantic segmentation: a review of methodologies and open challenges.
\newblock \emph{arXiv preprint arXiv:2304.05832}, 2023.

\bibitem[Chaslot et~al.(2008)Chaslot, Bakkes, Szita, and Spronck]{mcts}
Guillaume Chaslot, Sander Bakkes, Istvan Szita, and Pieter Spronck.
\newblock Monte-carlo tree search: A new framework for game ai.
\newblock In \emph{Proceedings of the AAAI Conference on Artificial Intelligence and Interactive Digital Entertainment}, volume~4, pages 216--217, 2008.

\bibitem[Chen et~al.(2017)Chen, Papandreou, Schroff, and Adam]{deeplabv3}
Liang-Chieh Chen, George Papandreou, Florian Schroff, and Hartwig Adam.
\newblock Rethinking atrous convolution for semantic image segmentation.
\newblock \emph{arXiv preprint arXiv:1706.05587}, 2017.

\bibitem[Chen et~al.(2019)Chen, Fu, Wang, Ma, Liu, and Hebert]{deform}
Zitian Chen, Yanwei Fu, Yu-Xiong Wang, Lin Ma, Wei Liu, and Martial Hebert.
\newblock Image deformation meta-networks for one-shot learning.
\newblock In \emph{Proceedings of the IEEE/CVF conference on computer vision and pattern recognition}, pages 8680--8689, 2019.

\bibitem[Chu et~al.(2019)Chu, Li, Chang, and Wang]{spot}
Wen-Hsuan Chu, Yu-Jhe Li, Jing-Cheng Chang, and Yu-Chiang~Frank Wang.
\newblock Spot and learn: A maximum-entropy patch sampler for few-shot image classification.
\newblock In \emph{Proceedings of the IEEE/CVF conference on computer vision and pattern recognition}, pages 6251--6260, 2019.

\bibitem[Crouse et~al.(2021)Crouse, Abdelaziz, Makni, Whitehead, Cornelio, Kapanipathi, Srinivas, Thost, Witbrock, and Fokoue]{proof-21}
Maxwell Crouse, Ibrahim Abdelaziz, Bassem Makni, Spencer Whitehead, Cristina Cornelio, Pavan Kapanipathi, Kavitha Srinivas, Veronika Thost, Michael Witbrock, and Achille Fokoue.
\newblock A deep reinforcement learning approach to first-order logic theorem proving.
\newblock In \emph{Proceedings of the AAAI Conference on Artificial Intelligence}, volume~35, pages 6279--6287, 2021.

\bibitem[Deng et~al.(2009)Deng, Dong, Socher, Li, Li, and Fei-Fei]{imagenet}
Jia Deng, Wei Dong, Richard Socher, Li-Jia Li, Kai Li, and Li~Fei-Fei.
\newblock Imagenet: A large-scale hierarchical image database.
\newblock In \emph{2009 IEEE Conference on Computer Vision and Pattern Recognition}, pages 248--255, 2009.
\newblock \doi{10.1109/CVPR.2009.5206848}.

\bibitem[Fawzi et~al.(2022)Fawzi, Balog, Huang, Hubert, Romera-Paredes, Barekatain, Novikov, R~Ruiz, Schrittwieser, Swirszcz, et~al.]{alphatensor}
Alhussein Fawzi, Matej Balog, Aja Huang, Thomas Hubert, Bernardino Romera-Paredes, Mohammadamin Barekatain, Alexander Novikov, Francisco~J R~Ruiz, Julian Schrittwieser, Grzegorz Swirszcz, et~al.
\newblock Discovering faster matrix multiplication algorithms with reinforcement learning.
\newblock \emph{Nature}, 610\penalty0 (7930):\penalty0 47--53, 2022.

\bibitem[Francis et~al.(2020)Francis, Mrziglod, Sidiropoulos, and Muller]{cloud}
Alistair Francis, John Mrziglod, Panagiotis Sidiropoulos, and Jan-Peter Muller.
\newblock Sentinel-2 cloud mask catalogue, 2020.
\newblock URL \url{https://zenodo.org/record/4172871}.

\bibitem[Ghorbanzadeh et~al.(2022)Ghorbanzadeh, Xu, Ghamisi, Kopp, and Kreil]{landslide}
Omid Ghorbanzadeh, Yonghao Xu, Pedram Ghamisi, Michael Kopp, and David Kreil.
\newblock Landslide4sense: Reference benchmark data and deep learning models for landslide detection.
\newblock \emph{IEEE Transactions on Geoscience and Remote Sensing}, 60:\penalty0 1--17, 2022.
\newblock \doi{10.1109/TGRS.2022.3215209}.

\bibitem[Ha et~al.(2017)Ha, Watanabe, Karasawa, Ushiku, and Harada]{mfnet}
Qishen Ha, Kohei Watanabe, Takumi Karasawa, Yoshitaka Ushiku, and Tatsuya Harada.
\newblock Mfnet: Towards real-time semantic segmentation for autonomous vehicles with multi-spectral scenes.
\newblock In \emph{2017 IEEE/RSJ International Conference on Intelligent Robots and Systems (IROS)}, pages 5108--5115. IEEE, 2017.

\bibitem[He et~al.(2016)He, Zhang, Ren, and Sun]{resnet}
Kaiming He, Xiangyu Zhang, Shaoqing Ren, and Jian Sun.
\newblock Deep residual learning for image recognition.
\newblock In \emph{Proceedings of the IEEE conference on computer vision and pattern recognition}, pages 770--778, 2016.

\bibitem[Hill et~al.(2003)Hill, Canagarajah, and Bull]{watershed}
Paul~R Hill, Cedric~Nishan Canagarajah, and David~R Bull.
\newblock Image segmentation using a texture gradient based watershed transform.
\newblock \emph{IEEE Transactions on Image Processing}, 12\penalty0 (12):\penalty0 1618--1633, 2003.

\bibitem[Kaliszyk et~al.(2018)Kaliszyk, Urban, Michalewski, and Ol{\v{s}}{\'a}k]{proof-18}
Cezary Kaliszyk, Josef Urban, Henryk Michalewski, and Miroslav Ol{\v{s}}{\'a}k.
\newblock Reinforcement learning of theorem proving.
\newblock \emph{Advances in Neural Information Processing Systems}, 31, 2018.

\bibitem[Kemker et~al.(2018)Kemker, Salvaggio, and Kanan]{rit-18}
Ronald Kemker, Carl Salvaggio, and Christopher Kanan.
\newblock Algorithms for semantic segmentation of multispectral remote sensing imagery using deep learning.
\newblock \emph{ISPRS Journal of Photogrammetry and Remote Sensing}, 2018.
\newblock ISSN 0924-2716.
\newblock \doi{https://doi.org/10.1016/j.isprsjprs.2018.04.014}.
\newblock URL \url{http://www.sciencedirect.com/science/article/pii/S0924271618301229}.

\bibitem[Landajuela et~al.(2021)Landajuela, Petersen, Kim, Santiago, Glatt, Mundhenk, Pettit, and Faissol]{symb-pol}
Mikel Landajuela, Brenden~K Petersen, Sookyung Kim, Claudio~P Santiago, Ruben Glatt, Nathan Mundhenk, Jacob~F Pettit, and Daniel Faissol.
\newblock Discovering symbolic policies with deep reinforcement learning.
\newblock In \emph{International Conference on Machine Learning}, pages 5979--5989. PMLR, 2021.

\bibitem[Loshchilov and Hutter(2017)]{adamw}
Ilya Loshchilov and Frank Hutter.
\newblock Fixing weight decay regularization in adam.
\newblock \emph{CoRR}, abs/1711.05101, 2017.
\newblock URL \url{http://arxiv.org/abs/1711.05101}.

\bibitem[Lv et~al.(2023)Lv, Liu, and Li]{cainet}
Ying Lv, Zhi Liu, and Gongyang Li.
\newblock Context-aware interaction network for rgb-t semantic segmentation.
\newblock \emph{IEEE Transactions on Multimedia}, pages 1--13, 2023.
\newblock \doi{10.1109/TMM.2023.3349072}.

\bibitem[Otsu(1979)]{otsu}
Nobuyuki Otsu.
\newblock A threshold selection method from gray-level histograms.
\newblock \emph{IEEE transactions on systems, man, and cybernetics}, 9\penalty0 (1):\penalty0 62--66, 1979.

\bibitem[Phadikar and Goswami(2016)]{riceseg}
Santanu Phadikar and Jyotirmoy Goswami.
\newblock Vegetation indices based segmentation for automatic classification of brown spot and blast diseases of rice.
\newblock In \emph{2016 3rd International Conference on Recent Advances in Information Technology (RAIT)}, pages 284--289. IEEE, 2016.

\bibitem[Ponti(2012)]{vegidxseg}
Moacir~P Ponti.
\newblock Segmentation of low-cost remote sensing images combining vegetation indices and mean shift.
\newblock \emph{IEEE Geoscience and Remote Sensing Letters}, 10\penalty0 (1):\penalty0 67--70, 2012.

\bibitem[Ronneberger et~al.(2015)Ronneberger, Fischer, and Brox]{unet}
Olaf Ronneberger, Philipp Fischer, and Thomas Brox.
\newblock U-net: Convolutional networks for biomedical image segmentation.
\newblock In \emph{Medical Image Computing and Computer-Assisted Intervention--MICCAI 2015: 18th International Conference, Munich, Germany, October 5-9, 2015, Proceedings, Part III 18}, pages 234--241. Springer, 2015.

\bibitem[Silver et~al.(2016)Silver, Huang, Maddison, Guez, Sifre, van~den Driessche, Schrittwieser, Antonoglou, Panneershelvam, Lanctot, Dieleman, Grewe, Nham, Kalchbrenner, Sutskever, Lillicrap, Leach, Kavukcuoglu, Graepel, and Hassabis]{alphago}
David Silver, Aja Huang, Chris~J. Maddison, Arthur Guez, Laurent Sifre, George van~den Driessche, Julian Schrittwieser, Ioannis Antonoglou, Veda Panneershelvam, Marc Lanctot, Sander Dieleman, Dominik Grewe, John Nham, Nal Kalchbrenner, Ilya Sutskever, Timothy Lillicrap, Madeleine Leach, Koray Kavukcuoglu, Thore Graepel, and Demis Hassabis.
\newblock Mastering the game of go with deep neural networks and tree search.
\newblock \emph{Nature}, 529\penalty0 (7587):\penalty0 484--489, January 2016.
\newblock \doi{10.1038/nature16961}.
\newblock URL \url{https://doi.org/10.1038/nature16961}.

\bibitem[Silver et~al.(2018)Silver, Hubert, Schrittwieser, Antonoglou, Lai, Guez, Lanctot, Sifre, Kumaran, Graepel, Lillicrap, Simonyan, and Hassabis]{alphazero}
David Silver, Thomas Hubert, Julian Schrittwieser, Ioannis Antonoglou, Matthew Lai, Arthur Guez, Marc Lanctot, Laurent Sifre, Dharshan Kumaran, Thore Graepel, Timothy Lillicrap, Karen Simonyan, and Demis Hassabis.
\newblock A general reinforcement learning algorithm that masters chess, shogi, and go through self-play.
\newblock \emph{Science}, 362\penalty0 (6419):\penalty0 1140--1144, 2018.
\newblock \doi{10.1126/science.aar6404}.
\newblock URL \url{https://www.science.org/doi/abs/10.1126/science.aar6404}.

\bibitem[Tao et~al.(2022)Tao, Meng, Li, Yang, Hu, Li, Cui, and Zhang]{msnet}
Chongxin Tao, Yizhuo Meng, Junjie Li, Beibei Yang, Fengmin Hu, Yuanxi Li, Changlu Cui, and Wen Zhang.
\newblock Msnet: multispectral semantic segmentation network for remote sensing images.
\newblock \emph{GIScience \& Remote Sensing}, 59\penalty0 (1):\penalty0 1177--1198, 2022.

\bibitem[Vayssade et~al.(2021)Vayssade, Paoli, G{\'e}e, and Jones]{deepindices}
Jehan-Antoine Vayssade, Jean-No{\"e}l Paoli, Christelle G{\'e}e, and Gawain Jones.
\newblock Deepindices: Remote sensing indices based on approximation of functions through deep-learning, application to uncalibrated vegetation images.
\newblock \emph{Remote Sensing}, 13\penalty0 (12):\penalty0 2261, 2021.

\bibitem[Vogiatzis and Stafylopatis(2000)]{symb-exp}
Dimitrios Vogiatzis and Andreas Stafylopatis.
\newblock Reinforcement learning for symbolic expression induction.
\newblock \emph{Mathematics and computers in simulation}, 51\penalty0 (3-4):\penalty0 169--179, 2000.

\bibitem[Wang et~al.(2022)Wang, Chen, Cheng, and Xiao]{vss}
Ke~Wang, Hainan Chen, Ligang Cheng, and Jian Xiao.
\newblock Variational-scale segmentation for multispectral remote-sensing images using spectral indices.
\newblock \emph{Remote Sensing}, 14\penalty0 (2):\penalty0 326, Jan 2022.
\newblock ISSN 2072-4292.
\newblock \doi{10.3390/rs14020326}.
\newblock URL \url{http://dx.doi.org/10.3390/rs14020326}.

\bibitem[Zhang et~al.(2019)Zhang, Zhang, and Koniusz]{hallu}
Hongguang Zhang, Jing Zhang, and Piotr Koniusz.
\newblock Few-shot learning via saliency-guided hallucination of samples.
\newblock In \emph{Proceedings of the IEEE/CVF Conference on Computer Vision and Pattern Recognition}, pages 2770--2779, 2019.

\bibitem[Zhao et~al.(2019)Zhao, Balakrishnan, Durand, Guttag, and Dalca]{med-aug}
Amy Zhao, Guha Balakrishnan, Fredo Durand, John~V Guttag, and Adrian~V Dalca.
\newblock Data augmentation using learned transformations for one-shot medical image segmentation.
\newblock In \emph{Proceedings of the IEEE/CVF conference on computer vision and pattern recognition}, pages 8543--8553, 2019.

\bibitem[Zhou et~al.(2018)Zhou, Rahman~Siddiquee, Tajbakhsh, and Liang]{unet++}
Zongwei Zhou, Md~Mahfuzur Rahman~Siddiquee, Nima Tajbakhsh, and Jianming Liang.
\newblock Unet++: A nested u-net architecture for medical image segmentation.
\newblock In \emph{Deep Learning in Medical Image Analysis and Multimodal Learning for Clinical Decision Support: 4th International Workshop, DLMIA 2018, and 8th International Workshop, ML-CDS 2018, Held in Conjunction with MICCAI 2018, Granada, Spain, September 20, 2018, Proceedings 4}, pages 3--11. Springer, 2018.

\end{thebibliography}
\end{document}

\maketitle

%-------------------------------------------------------------------------

\appendix

% \begin{figure}
%   \includegraphics[width=\textwidth]{figures/Index Generator.pdf}
%   \caption{\textbf{Index Generator Training Process.} Each training iteration of the index generator has 2 main phases: the data collection phase and the policy training phase. Expression samples are generated during the data collection phase by a generative model that guides MCTS. The model is improved by using the generated samples to train the policy network during the policy training phase.}
%   \label{fig:index-generator}
%   \centering
% \end{figure}

\section{Implementation Details}

\subsection{Adaptations}
\label{sec:adaptations}

\paragraph{Pretraining for valid output generation.}
\label{sec:pretraining}
Each iteration of MCTS spends a non-negligible amount of time evaluating the expression. Since the expression validity can be evaluated with a $(1,1,n_{channels})$-shaped tensor, the expression evaluation time can be significantly reduced during the pretraining stage. As a result, the model can learn to generate valid expressions much faster. This also gives the added benefit of providing a pretrained model to initialize weights for a new task, assuming the new task works with the same number of channels.

During the pretraining phase, certain tendencies are observed in the behavior of the agent. Firstly, the agent often generates short expressions. This may be because the more complex the expression, the easier it is to deviate from a regular range of values. Secondly, the agent tends to avoid opening parentheses. This is likely caused by the fact that once an opening parenthesis is generated, the expression remains invalid until a closing parenthesis is generated and this leads to a higher chance of negative rewards.

It is also observed that a significant fraction of existing remote-sensing indices is "unitless". In other words, if each channel of the multispectral image is given a unit of measurement, the spectral index resulting from a mathematical operation between those bands lacks a unit.

Accordingly, to motivate the agent to address the aforementioned considerations, the reward for pretraining $r(s)$ is defined as presented in equation \ref{eq:improved-rew}.

\begin{eqnarray}
r_{len} & = & 0.02 \times l_{exp}\\
r_{par} & = & 0.2 \times n_{par}\\
r_{unit} & = & \begin{cases}
    1 \text{, if expression is unitless}\\
    0 \text{, otherwise}\\
\end{cases} \\
r(s) & = & 0.5 + r_{len} + r_{par} + r_{unit} \label{eq:improved-rew}
\end{eqnarray}

where $l_{exp}$ is the length of the expression and $n_{par}$ is the number of pairs of parentheses in the expression.

\paragraph{Action validity.}

\begin{table}
  \centering
  \begin{tabular}{ll}
    \hline
    Previous Action         & Valid Actions      \\
    \hline
    $start, (, +, -, \times, /$    & $(,channel$         \\
    $channel, ) $             & $+, -, \times, /, ), =$  \\
    \hline
  \end{tabular}
  \caption{The list of valid actions, given the previous action, where $start$ refers to the starting state (empty expression) and $channel$ refers to a channel of the image.}
  \label{tab:action-validity-table}
\end{table}

At each state, we define a set of valid actions based on the previous action as shown in Table \ref{tab:action-validity-table}. In addition, certain other checks are performed to avoid invalidity and redundancy where possible.

\begin{itemize}
    \item The number of “)” symbols in the expression is always maintained to be less than or equal to the number of “(“ symbols in the expression. In other words, generating the “)” symbol is defined to be invalid if all opened parentheses have already been closed.
    \item Since enclosing a single symbol within parentheses is redundant as the parentheses can simply be removed, generating a closing parenthesis, two actions after generating an opening parenthesis, is defined to be an invalid action.
\end{itemize}

\paragraph{Adaptive Data Buffer.}
\label{section:adaptive-buffer}

The classification of expressions as high-reward or low-reward can be achieved by various criteria. For the experiments performed in this research, this classification is performed by using an adaptive data buffer. The idea is to progressively reduce the size of the data buffer to get the GPT-based model to overfit to a set of expressions that provide a higher reward. We chose to implement this functionality as follows. The data collection phase is initially executed until the buffer reaches a certain capacity. The iterations that follow execute both the data collection phase and training phase. After each iteration, the buffer size is set to 95\% of its current capacity, dropping the expressions within the 5\% of lowest rewards. This is repeated until a certain minimum capacity is reached. This minimum capacity would be a relatively low number (e.g.: 20), such that the model may overfit to those expressions while also exploring expressions that contain similar symbols and structures.

% \paragraph{Dynamic MCTS iteration counts.}
% Each data collection episode begins with 1000 iterations of exploration for making the first action and decreases the number of iterations by 50 for each subsequent step. This is because having too many iterations near the end of the episode may be wasted computation.

% \paragraph{Increased probability for generating the “(“ symbol.}
% It was observed that the agent is hesitant to generate the “(“ symbol since this would result in the expression being in an invalid state until the “)” symbol is generated. As a workaround, the value of picking the “(“ symbol was multiplied by some constant (>1) to increase the assigned softmax value in the action probability distribution. This increase is maintained for several pretraining epochs and disabled during fine-tuning.

\subsection{Evaluated Index Preprocessing}
\label{sec:index-proc}

Prior to being used in reward calculation, the evaluated index is updated by standardizing, clipping, and scaling to a [0, 1] range (Eq. \ref{eq:standardize} - \ref{eq:scale}).

\begin{align}
    standardize(\mathcal{E}) & = \frac{\mathcal{E} - \mu_\mathcal{E}}{\sigma_\mathcal{E}} \label{eq:standardize} \\
    clip(\mathcal{E}) & = max\{min\{\mathcal{E}, Z_{max}\}, Z_{min}\} \label{eq:clip} \\
    scale(\mathcal{E}) & = \frac{\mathcal{E} - Z_{min}}{Z_{max} - Z_{min}} \label{eq:scale}
\end{align}

\noindent
where $\mathcal{E}$ is the evaluated index, $\mu_\mathcal{E}$ and $\sigma_\mathcal{E}$ are the channel-wise means and standard deviations of $\mathcal{E}$, and $Z_{min}$ and $Z_{max}$ are the minimum and maximum Z-scores permitted. We use $Z_{min} = -3$ and $Z_{max} = 3$ in our experiments.

\subsection{GPT-based Model Configurations}
\label{sec:gpt-config}
\subsubsection{Model Architecture}
Number of layers = $4$ \\
Number of attention heads = $4$ \\
Embedding size = $128$ \\
Dropout = $0.0$

\subsubsection{Adam Optimizer}
Learning rate = $1e-4$ \\
Weight decay = $0.1$ \\
Beta1 = $0.9$ \\
Beta2 = $0.95$

\begin{table}
  \centering
  \begin{tabular}{llll}
    \hline
    Function & Correlation & t-statistic & p-value     \\
    \hline
    IoU & 0.0554 & 0.7807 & 0.4359 \\
    CS & 0.0599 & 0.8444  & 0.3995 \\
    F1 & 0.0667 & 0.9406 & 0.3481 \\
    AUC & 0.0776 & 1.0952 &  0.2748  \\
    PCC & \textbf{0.1417} & \textbf{2.0142} & \textbf{0.0453} \\
    \hline
  \end{tabular}
  \caption{A statistical comparison of the correlations of each heuristic function with the training score}
  \label{tab:heuristic-comparison}
\end{table}

\section{Additional Experiments}
\subsection{Qualitative Comparison}
\label{sec:qualitative}

\begin{figure*}[ht]
  \centering
  \includegraphics[width=\linewidth]{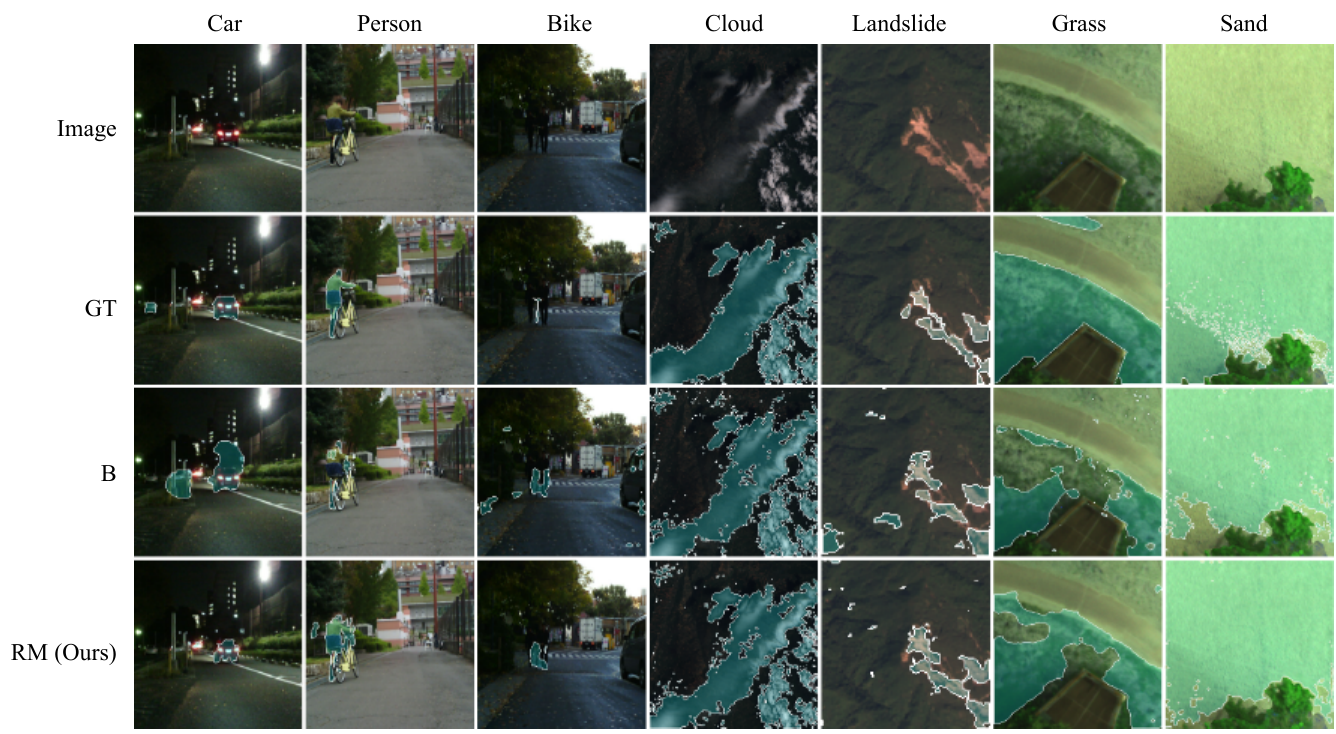}
  \caption{Qualitative Comparison between the baseline result (B) and Multi-index replacement (RM) across the datasets with the UNet model, along with the ground truth (GT).}
  \label{fig:qualitative}
\end{figure*}

We perform a qualitative comparison between the baseline (B) model and the multi-index replacement (RM) mode (Figure \ref{fig:qualitative}). Through the comparison, we observe that the RM mode tends to be relatively less confused by objects that blend into the background in terms of color.

\subsection{Effects of Generated Indices versus Existing Indices.}
\label{sec:ndvi}

\begin{table}
  \centering
  \begin{tabular}{l|l|ll|ll}
    \hline
    \multirow{2}{*}{Size}   & \multirow{2}{*}{Baseline}   &  \multicolumn{2}{c|}{NDVI} & \multicolumn{2}{c}{Generated}   \\
    \cline{3-6}
                           & & R & RM & R & RM \\
    \hline
    UNet    & 58.0 & 74.0 & 73.6 & 73.6 & 73.7 \\
    UNet++  & 60.9 & 70.2 & 72.4 & 73.6 & 70.3 \\
    \hline
  \end{tabular}
  \caption{\textbf{Effects of Generated Indices versus Existing Indices}. The table shows the comparison between the IoU scores of the multiple-index replacement method (RM) and the single-index replacement method (R) for the NDVI index and the index generated by the proposed method for the Grass class of the RIT-18 dataset.}
  \label{tab:ndvi}
\end{table}

To evaluate the performance of the generated indices in comparison to pre-existing indices, we compare NDVI to the expression for the Grass class of the RIT-18 dataset. We choose this setting due to the specific use of NDVI in the identification of greenery in prior work. Despite the NDVI index being specifically created for this context, we observer comparable results when using the generated indices

\section{Code}
Please note that the codebase for the project is available in the supplementary material in the \texttt{code.zip} file. The \texttt{README.md} file within the codebase provides instructions on how to set up the workspace, download the dataset, and execute the algorithms.

Once the dataset is downloaded and extracted, the resulting directory structure is expected to look as follows:
\begin{verbatim}
./
|- dataset/
|- indexrl/
|_ dataset.py
|_ dataset_config.py
|_ ...

\end{verbatim}

% \bibliography{egbib}